\documentclass[10pt,twocolumn,letterpaper]{article}

\usepackage{cvpr}
\usepackage{times}
\usepackage{epsfig}
\usepackage{enumerate}
\usepackage{graphicx}
\usepackage{amsmath}
\usepackage{amssymb}
\usepackage{diagbox}
\usepackage{multirow}
\usepackage{subfigure}
\usepackage{color}
\usepackage{verbatim}
\usepackage{array}
\usepackage{url}
\usepackage{bm}
\usepackage[pagebackref=true,breaklinks=true,letterpaper=true,colorlinks,bookmarks=false]{hyperref}
\cvprfinalcopy 

\def\cvprPaperID{229} 

\ifcvprfinal\fi
\begin{document}
\hyphenpenalty=400

\title{Face Alignment Across Large Poses: A 3D Solution}

\author{Xiangyu Zhu$^{1}$ \and Zhen Lei$^{1}$ \and Xiaoming Liu$^{2}$ \and Hailin Shi$^{1}$ \and Stan Z. Li$^{1}$ \and
$^{1}$Institute of Automation, Chinese Academy of Sciences \\
$^{2}$Department of Computer Science and Engineering, Michigan State University\\
{\tt \small \{xiangyu.zhu,zlei,hailin.shi,szli\}@nlpr.ia.ac.cn} ~~~~~~~ {\tt \small liuxm@msu.edu }}

\maketitle

\begin{abstract}
Face alignment, which fits a face model to an image and extracts the semantic meanings of facial pixels, has been an important topic in CV community. However, most algorithms are designed for faces in small to medium poses (below $45^{\circ}$), lacking the ability to align faces in large poses up to $90^{\circ}$. The challenges are three-fold: Firstly, the commonly used landmark-based face model assumes that all the landmarks are visible and is therefore not suitable for profile views. Secondly, the face appearance varies more dramatically across large poses, ranging from frontal view to profile view. Thirdly, labelling landmarks in large poses is extremely challenging since the invisible landmarks have to be guessed. In this paper, we propose a solution to the three problems in an new alignment framework, called 3D Dense Face Alignment (3DDFA), in which a dense 3D face model is fitted to the image via convolutional neutral network (CNN). We also propose a method to synthesize large-scale training samples in profile views to solve the third problem of data labelling. Experiments on the challenging AFLW database show that our approach achieves significant improvements over state-of-the-art methods.
\end{abstract}

\section{Introduction}\label{sec-introduction}
Traditional face alignment aims to locate face fiducial points like ``eye corner'', ``nose tip'' and ``chin center'', based on which the face image can be normalized. It is an essential preprocessing step for many face analysis tasks, e.g., face recognition~\cite{Taigman-CVPR-2013}, expression recognition~\cite{bettadapura2012face} and inverse rendering~\cite{Aldrian-PAMI-13}. The researches in face alignment can be divided into two categories: the analysis-by-synthesis based~\cite{Cootes-ECCV-98,tzimiropoulos-ICCV-13,Cristinacce-PR-08} and regression based~\cite{cootes1998comparative,Dollar-CVPR-10,lee2015face,Xiong-CVPR-13}. The former simulates the process of image generation and achieves alignment by minimizing the difference between model appearance and input image. The latter extracts features around key points and regresses it to the ground truth landmarks. With the development in the last decade, face alignment across medium poses, where the yaw angle is less than $45^{\circ}$ and all the landmarks are visible, has been well addressed~\cite{Xiong-CVPR-13,zhang2014facial,zhu2015face}.
However, face alignment across large poses ($\pm 90^{\circ}$) is still a challenging problem without much attention and achievements. There are three main challenges:

\begin{figure}[!htb]
  \centering
  \includegraphics[width=0.45\textwidth]{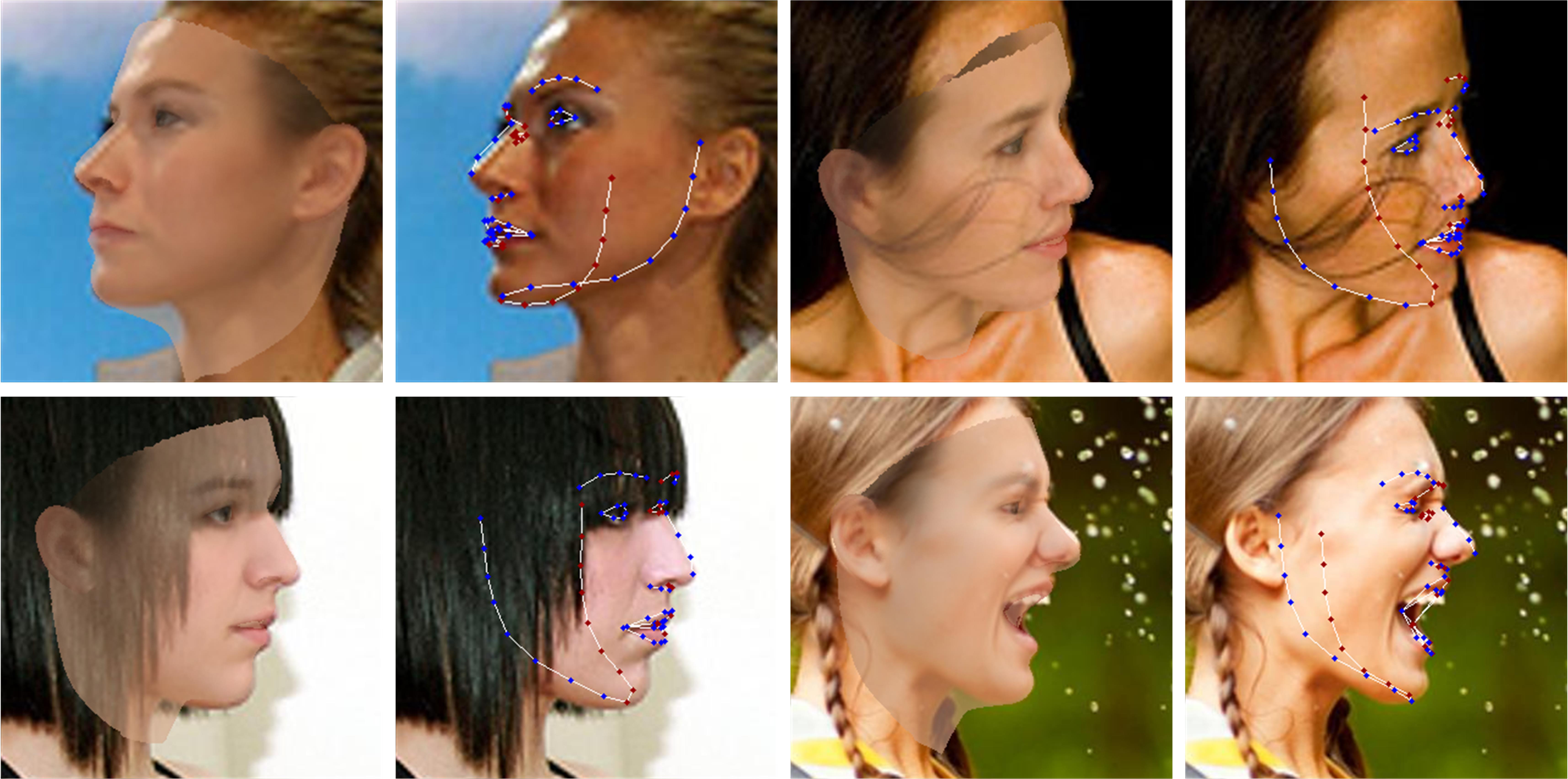}
  \caption{Fitting results of 3DDFA. For each pair of the four results, on the left is the rendering of the fitted 3D shape with the mean texture, which is made transparent to demonstrate the fitting accuracy. On the right is the landmarks overlayed on the 3D face model, in which the blue/red ones indicate visible/invisible landmarks. The visibility is directly computed from the fitted dense model by~\cite{hassner2014effective}. More results are demonstrated in supplemental material.}
  \label{fig-demo}
\end{figure}

\textbf{Modelling}: Landmark shape model~\cite{cootes1995active} implicitly assumes that each landmark can be robustly detected based on its distinctive visual patterns. However, when faces deviate from the frontal view, some landmarks become invisible due to self-occlusion~\cite{zhou2005bayesian}. In medium poses, this problem can be addressed by changing the semantic positions of face contour landmarks to the silhouette, which is termed landmark marching~\cite{zhu2015high}. However, in large poses where half of face is occluded, some landmarks are inevitably invisible and have no image data. As a result, the landmark shape model no longer works well.

\textbf{Fitting}: Face alignment across large poses is more challenging than medium poses due to the dramatic appearance variations when close to the profile views. The cascaded linear regression~\cite{Xiong-CVPR-13} or traditional nonlinear models~\cite{lee2015face,zhang2014coarse,Cao-CVPR-12} are not sophisticated enough to cover such complicated patterns in a unified way. The view-based framework, which adopts different landmark configurations and fitting models for each view category~\cite{zhou2005bayesian,yu2013pose,zhu2012face,smith2014nonparametric}, may significantly increase computation cost since every view has to be tested.

\textbf{Data Labelling}: The most serious problem comes from the data. Manual labelling landmarks on large-pose faces is a very tedious task. Firstly, no algorithm can provide a good initialization to reduce the workload. Secondly, the occluded landmarks have to be ``guessed'' which is impossible for most of people. As a result, almost all public face alignment databases such as AFW~\cite{zhu2012face}, LFPW~\cite{jaiswal2013guided}, HELEN~\cite{le2012interactive} and IBUG~\cite{sagonas2013semi} are collected in medium poses. Existing large-pose databases such as AFLW~\cite{kostinger2011annotated} only contains visible landmarks, which could be ambiguous in invisible landmarks and hard to train a unified face alignment model.

In this paper, we address all the three challenges with the goal of improving the face alignment performance across large poses.

\begin{enumerate}
\item To address the problem of invisible landmarks in large poses, we propose to fit the 3D dense face model rather than the sparse landmark shape model to the image. By incorporating 3D information, the appearance variations and self-occlusion caused by 3D transformations can be inherently addressed. We call this method 3D Dense Face Alignment (3DDFA). Some results are shown in Fig.~\ref{fig-demo}.
\item To resolve the fitting process in 3DDFA, we propose a cascaded convolutional neutral network (CNN) based regression method. CNN has been proved of excellent capability to extract useful information from images with large variations in object detection~\cite{yan2015object} and image classification~\cite{szegedy2014going}. In this work, we adopt CNN to fit the 3D face model with a specifically designed feature, namely Projected Normalized Coordinate Code (PNCC). Besides, Weighted Parameter Distance Cost (WPDC) is proposed as the cost function. To the best of our knowledge, this is the first attempt to solve the 3D face alignment with CNN.
\item To enable the training of the 3DDFA, we construct a face database containing pairs of 2D face images and 3D face models. We further propose a face profiling algorithm to synthesize $60k+$ training samples across large poses. The synthesized samples well simulate the face appearances in large poses and boost the performance of both prior and our proposed face alignment algorithms.
\end{enumerate}

The database, face profiling code and 3DDFA code are released at \url{http://www.cbsr.ia.ac.cn/users/xiangyuzhu/}.

\section{Related Works}

\textbf{Generic Face Alignment}: Face alignment in 2D aims at locating a sparse set of fiducial facial landmarks. A number of achievements have been made including the classic Active Appearance Model (AAM)~\cite{Cootes-ECCV-98,Saragih-ICCV-07,tzimiropoulos-ICCV-13} and Constrained Local Model (CLM)~\cite{Cristinacce-BMVC-06,Saragih-IJCV-10,Asthana-CVPR-13}. Recently, the regression based method, which maps the discriminative features around landmarks to the desired landmark positions~\cite{Valstar-CVPR-10,Xiong-CVPR-13,xiong2015global,Cao-CVPR-12,zhang2014coarse,lee2015face}, has been proposed. By utilizing the feedback characteristic that the the output (landmark positions) of the regression has an influence on the input (features at landmarks), the cascaded regression~\cite{Dollar-CVPR-10} cascades a list of weak regressors to reduce the alignment error progressively and reaches the state of the art~\cite{xiong2015global,zhu2015face}.

Besides traditional models, convolutional neutral network (CNN) has also been employed in face alignment recently. Sun et al.~\cite{sun2013deep} firstly use CNN to regress landmark locations with the raw face image. Liang et al.~\cite{liang2015unconstrained} improve the flexibility by estimating the landmark response map. Zhang et al.~\cite{zhang2014facial} further combine face alignment with attribute analysis through multi-task CNN to boost the performance of both tasks. Although with considerable achievements, most CNN methods only detect a sparse set of landmarks (5 points in~\cite{sun2013deep,zhang2014facial,liang2015unconstrained}) with limited descriptive power of face shape.

\textbf{Large Pose Face Alignment}: Despite the great attentions on face alignment, literature on large-pose scenario is rather limited. The most common method is the multi-view framework~\cite{cootes2002view}, which uses different landmark configurations for different views. For example, TSPM~\cite{zhu2012face} and CDM~\cite{yu2013pose} employ DPM-like~\cite{felzenszwalb2010object} method to align faces with different shape models, among which the highest possibility is chosen as the final result. However, since every view has to be tested, the computation cost of multi-view method is always high.

Besides 2D methods, 3D face alignment~\cite{gu20063D}, which aims to fit a 3D morphable model (3DMM)~\cite{Blanz-PAMI-03} from a 2D image, also has the potential to deal with large poses. It models the 3D face shape with a linear subspace (PCA~\cite{Blanz-PAMI-03} or Tensor~\cite{Cao-2014-SIG}) and achieves fitting by minimizing the difference between image and model appearance. 3DMM can cover arbitrary poses~\cite{Blanz-PAMI-03,Romdhani-CVPR-05} but suffers from the one-minute-per-image computation cost. Recently, regression based 3DMM fitting, which estimates the model parameters by regressing the features at landmark positions~\cite{yu2013pose,jourabloo2015pose,Cao-2014-SIG,jeni2015dense}, has been proposed to improve the efficiency. However, since the features at landmarks may be self-occluded as in 2D methods, the fitting algorithm is no longer pose-invariant and suffers from the three problems in Section~\ref{sec-introduction}. A relevant but different problem is the 3D face reconstruction~\cite{Aldrian-PAMI-13,xiao2004real,zhu2015high,hassner2013viewing}, which recovers a 3D face from given 2D landmarks. Interestingly, based on that 2D/3D face alignment results can be mutually transformed, where 3D to 2D is made by selecting $x,y$ coordinates of landmark vertexes and 2D to 3D is made by 3D face reconstruction.

\begin{figure*}
  \centering
  \includegraphics[width=0.95\textwidth]{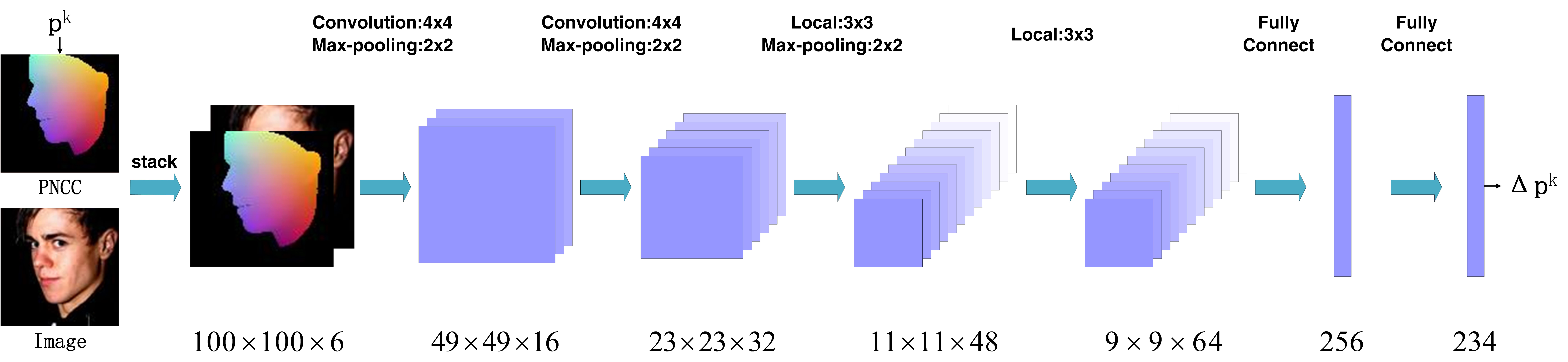}
  \caption{An overview of 3DDFA. At $k$th iteration, $\emph{Net}^{k}$ takes a medium parameter $\mathbf{p}^{k}$ as input, constructs the projected normalized coordinate code (PNCC), stacks it with the input image and sends it into CNN to predict the parameter update $\Delta \mathbf{p}^{k}$.}
  \label{fig-overview}
\end{figure*}

\section{3D Dense Face Alignment (3DDFA)}
In this section we introduce the 3D Dense Face Alignment (3DDFA) which fits 3D morphable model with cascaded CNN.

\subsection{3D Morphable Model}
Blanz et al.~\cite{Blanz-PAMI-03} propose the 3D morphable model (3DMM) which describes the 3D face space with PCA:
\begin{equation}\label{equ-tensor}
  \mathbf{S}=\mathbf{\overline{S}} + \mathbf{A}_{id}\bm{\alpha}_{id} + \mathbf{A}_{exp}\bm{\alpha}_{exp},
\end{equation}
where $\mathbf{S}$ is a 3D face, $\mathbf{\overline{S}}$ is the mean shape, $\mathbf{A}_{id}$ is the principle axes trained on the 3D face scans with neutral expression and $\bm{\alpha}_{id}$ is the shape parameter, $\mathbf{A}_{exp}$ is the principle axes trained on the offsets between expression scans and neutral scans and $\bm{\alpha}_{exp}$ is the expression parameter. In this work, the $\mathbf{A}_{id}$ and $\mathbf{A}_{exp}$ come from BFM~\cite{Paysan-AVSS-09} and FaceWarehouse~\cite{Cao-2013-Facewarehouse} respectively. The 3D face is then projected onto the image plane with Weak Perspective Projection:
\begin{equation}\label{equ-projection}
  V(\mathbf{p}) = f*\mathbf{Pr} * \mathbf{R}*(\mathbf{\overline{S}} + \mathbf{A}_{id}\bm{\alpha}_{id} + \mathbf{A}_{exp}\bm{\alpha}_{exp}) +\mathbf{t}_{2d},
\end{equation}
where $V(\mathbf{p})$ is the model construction and projection function, leading to the 2D positions of model vertexes, $f$ is the scale factor, $\mathbf{Pr}$ is the orthographic projection matrix
$\left(
\begin{array}{ccc}
1 & 0 & 0 \\
0 & 1 & 0 \\
\end{array}
\right)
$
, $\mathbf{R}$ is the rotation matrix constructed from rotation angles $pitch$, $yaw$, $roll$ and $\mathbf{t}_{2d}$ is the translation vector. The collection of all the model parameters is $\mathbf{p}=[f,pitch,yaw,roll,\mathbf{t}_{2d},\bm{\alpha}_{id},\bm{\alpha}_{exp}]^{T}$.

\subsection{Network Structure}
The purpose of 3D face alignment is estimating $\mathbf{p}$ from a single face image $\mathbf{I}$. Unlike existing CNN methods~\cite{sun2013deep,liang2015unconstrained} which apply different networks for different fitting stages, 3DDFA employ a unified network structure across the cascade. In general, at iteration $k$ ($k=0,1,...,K$), given an initial parameter $\mathbf{p}^{k}$, we construct a specially designed feature PNCC with $\mathbf{p}^{k}$ and train a convolutional neutral network $\emph{Net}^{k}$ to predict the parameter update $\Delta \mathbf{p}^{k}$:
\begin{equation}\label{equ-3DDFA}
  \Delta \mathbf{p}^{k} = \emph{Net}^{k} (\mathbf{I}, \text{PNCC}(\mathbf{p}^{k})).
\end{equation}
Afterwards, a better medium parameter $\mathbf{p}^{k+1}=\mathbf{p}^{k}+\Delta \mathbf{p}^{k}$ becomes the input of the next network $\emph{Net}^{k+1}$ which has the same structure as $\emph{Net}^{k}$.
Fig.~\ref{fig-overview} shows the network structure. The input is the $100\times100\times3$ color image stacked by PNCC. The network contains four convolution layers, three pooling layers and two fully connected layers. The first two convolution layers share weights to extract low-level features. The last two convolution layers do not share weights to extract location sensitive features, which is further regressed to a 256-dimensional feature vector. The output is a 234-dimensional parameter update including 6-dimensional pose parameters $[f, pitch, yaw, roll, t_{2dx}, t_{2dy}]$, 199-dimensional shape parameters $\bm{\alpha}_{id}$ and 29-dimensional expression parameters $\bm{\alpha}_{exp}$.

\subsection{Projected Normalized Coordinate Code}
The special structure of the cascaded CNN has three requirements of its input feature: Firstly, the \textbf{feedback property} requires that the input feature should depend on the CNN output to enable the cascade manner. Secondly, the \textbf{convergence property} requires that the input feature should reflect the fitting accuracy to make the cascade converge after some iterations~\cite{zhu2015discriminative}. Finally, the \textbf{convolvable property} requires that the convolution on the input feature should make sense.
\begin{figure}[!htb]
  \centering
  \subfigure[NCC]{\label{fig-PNCC-a}
  \includegraphics[width=0.225\textwidth]{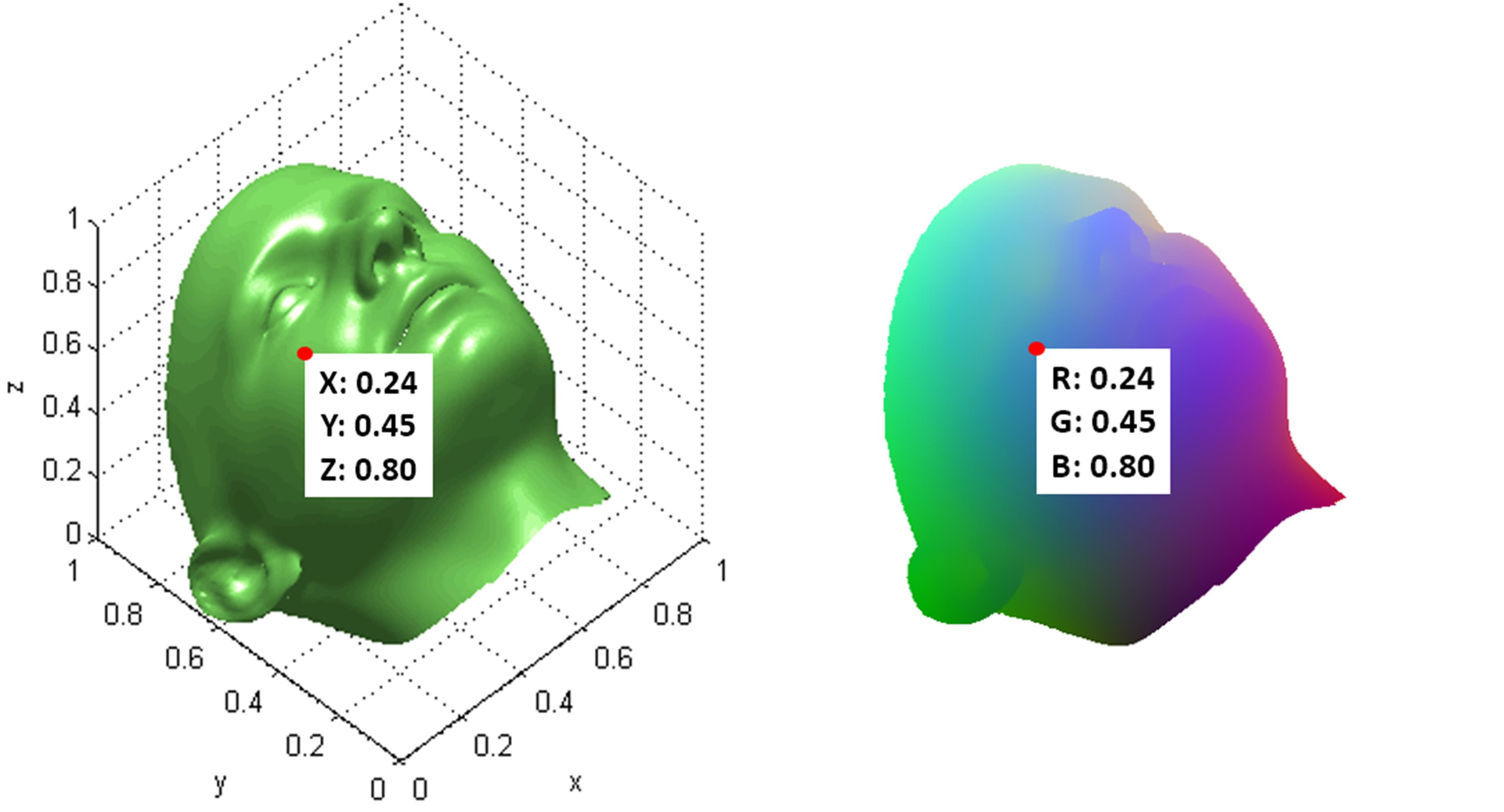}}
  \subfigure[PNCC]{\label{fig-PNCC-b}
  \includegraphics[width=0.225\textwidth]{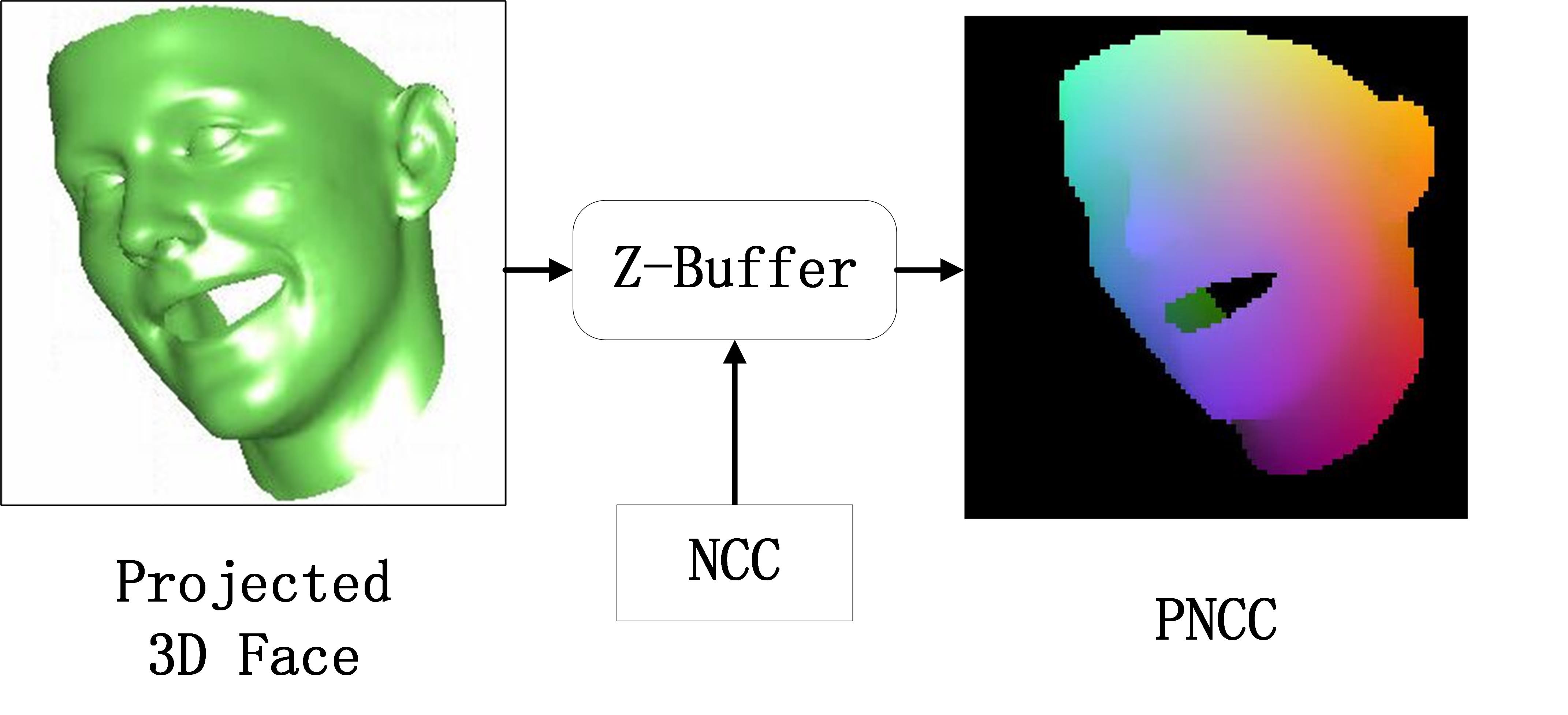}}
  \caption{The Normalized Coordinate Code (NCC) and the Projected Normalized Coordinate Code (PNCC). (a) The normalized mean face, which is also demonstrated with NCC as its texture ($\text{NCC}_{x}=\text{R}$, $\text{NCC}_{y}=\text{G}$, $\text{NCC}_{z}=\text{B}$). (b) The generation of PNCC: The projected 3D face is rendered by Z-Buffer with NCC as its colormap.}
  \label{fig-PNCC}
\end{figure}
Based on the three properties, we design our features as follows: Firstly, the 3D mean face is normalized to $0-1$ in $x,y,z$ axis as Equ.~\ref{equ-NCC}. The unique 3D coordinate of each vertex is called its Normalized Coordinate Code (NCC), see Fig.~\ref{fig-PNCC-a}.
\begin{equation}\label{equ-NCC}
  \text{NCC}_{d}=\frac{\mathbf{\overline{S}}_{d} - \min(\overline{\mathbf{S}}_{d})}{\max(\overline{\mathbf{S}}_{d}) - \min(\overline{\mathbf{S}}_{d})}~~~ (d = x,y,z),
\end{equation}
where the $\mathbf{\overline{S}}$ is the mean shape of 3DMM in Equ.~\ref{equ-tensor}. Since NCC has three channels as RGB, we also show the mean face with NCC as its texture. Secondly, with a model parameter $\mathbf{p}$, we adopt the Z-Buffer to render the projected 3D face colored by NCC as in Equ.~\ref{equ-PNCC}, which is called the Projected Normalized Coordinate Code (PNCC), see Fig.~\ref{fig-PNCC-b}:
\begin{gather}
   \text{PNCC}= \emph{Z-Buffer}(V_{3d}(\mathbf{p}),  \text{NCC}) \notag\\
  V_{3d}(\mathbf{p}) = f * \mathbf{R} * \mathbf{S} + [\mathbf{t}_{2d},0]^{T} \label{equ-PNCC}\\
  \mathbf{S} = \mathbf{\overline{S}} + \mathbf{A}_{id}\bm{\alpha}_{id} + \mathbf{A}_{exp}\bm{\alpha}_{exp}, \notag
\end{gather}
where $\emph{Z-Buffer}(\bm{\nu},\bm{\tau})$ renders an image from the 3D mesh $\bm{\nu}$ colored by $\bm{\tau}$ and $V_{3d}(\mathbf{p})$ is the current 3D face. Afterwards, PNCC is stacked with the input image and transferred to CNN.
Regarding the three properties, PNCC fulfills the feedback property since in Equ.~\ref{equ-PNCC}, $\mathbf{p}$ is the output of CNN and NCC is a constant. Secondly, PNCC provides the 2D locations of visible 3D vertexes on the image plane. When CNN detects that each NCC superposes its corresponding image pattern during testing, the cascade will converge. PNCC fulfills the convergence property. Note that the invisible region is automatically ignored by Z-Buffer. Finally, PNCC is smooth in 2D space, the convolution indicates the linear combination of NCCs on a local patch. It fulfills the convolvable property.

\subsection{Cost Function}
The performance of CNN can be greatly impacted by the cost function, which is difficult to design in 3DDFA since each dimension of the CNN output (model parameter) has different influence on the 3DDFA result (fitted 3D face). In this work, we discuss two baselines and propose a novel cost function. Since the parameter range varies significantly, we conduct z-score normalization before training.

\subsubsection{Parameter Distance Cost (PDC)}
Take the first iteration as an example. The purpose of CNN is predicting the parameter update $\Delta \mathbf{p}$ to move the initial parameter $\mathbf{p}^{0}$ closer to the ground truth $\mathbf{p}^{g}$.
Intuitively, we can minimize the distance between the ground truth and the current parameter with the Parameter Distance Cost (PDC):
\begin{equation}\label{equ-pdc}
  E_{pdc} = \| \Delta \mathbf{p} - (\mathbf{p}^{g}-\mathbf{p}^{0}) \|^{2}.
\end{equation}
Even though PDC has been used in 3D face alignment~\cite{zhu2015discriminative}, there is a problem that each dimension in $\mathbf{p}$ has different influence on the resultant 3D face. For example, with the same deviation, the yaw angle will bring a larger alignment error than a shape PCA coefficient, while PDC optimizes them equally.

\subsubsection{Vertex Distance Cost (VDC)}
Since 3DDFA aims to morph the 3DMM to the ground truth 3D face, we can optimize $\Delta \mathbf{p}$ by minimizing the vertex distances between the fitted and the ground truth 3D face:
\begin{equation}\label{equ-vdc}
  E_{vdc} = \| V(\mathbf{p}^{0} + \Delta \mathbf{p}) - V(\mathbf{p}^{g}) \|^{2},
\end{equation}
where $V(\cdot)$ is the face construction and weak perspective projection as Equ.~\ref{equ-projection}. This cost is called the Vertex Distance Cost (VDC) and
the derivative is provided in supplemental material. Compared with PDC, VDC better models the fitting error by explicitly considering the semantics of each parameter. However, we observe that VDC exhibits pathological curvature~\cite{martens2010deep}. The directions of pose parameters always exhibit much higher curvatures than the PCA coefficients. As a result, optimizing VDC with gradient descend converges very slowly due to the ``zig-zagging'' problem. Second-order optimizations are preferred but they are expensive and hard to be implemented on GPU.

\subsubsection{Weighted Parameter Distance Cost (WPDC)}
In this work, we propose a simple but effective cost function Weighted Parameter Distance Cost (WPDC). The basic idea is explicitly modeling the importance of each parameter:
\begin{equation}
\begin{split}
  E_{wpdc} = (\Delta \mathbf{p} - (\mathbf{p}^{g}-\mathbf{p}^{0}&))^{T}\mathbf{W}(\Delta \mathbf{p} - (\mathbf{p}^{g}-\mathbf{p}^{0})) \\
  where~~~~~~~\mathbf{W} = diag(w_{1},&w_{2},...,w_{n})\\
            w_{i} = \| V(\mathbf{p}^{d}(i)) -& V(\mathbf{p}^{g}) \| / \sum w_{i}  \\
            \mathbf{p}^{d}(i)_{i} = (\mathbf{p}^{0}&+\Delta \mathbf{p})_{i}\\
            \mathbf{p}^{d}(i)_{j} = \mathbf{p}^{g}_{j},~~~ j\in{} \{1,&\ldots,i-1,i+1,\ldots,n\},\\
\end{split}
\label{equ-wpdc}
\end{equation}
where $\mathbf{W}$ is the importance matrix whose diagonal is the weight of each parameter, $\mathbf{p}^{d}(i)$ is the i-deteriorated parameter whose $i$th component comes from the predicted parameter $(\mathbf{p}^{0}+\Delta \mathbf{p})$ and the others come from the ground truth parameter $\mathbf{p}^{g}$, $\| V(\mathbf{p}^{d}(i)) - V(\mathbf{p}^{g}) \|$ models the alignment error brought by miss-predicting the $i$th model parameter, which is indicative of its importance. For simplicity, $\mathbf{W}$ is considered as a constant when computing the derivative. In the training process, CNN firstly concentrate on the parameters with larger $\| V(\mathbf{p}^{d}(i)) - V(\mathbf{p}^{g}) \|$ such as scale, rotation and translation. As $\mathbf{p}^{d}(i)$ is closer to $\mathbf{p}^{g}$, the weights of these parameters begin to shrink and CNN will optimize less important parameters but at the same time keep the high-priority parameters sufficiently good. Compared with VDC, the WPDC remedies the pathological curvature issue and is easier to implement without the derivative of $V(\cdot)$.

\section{Face Profiling}
All the discriminative models rely on the training data, especially for CNN which has thousands of parameters to train. Therefore, massive labelled faces across large poses are crucial for 3DDFA. However, few of released face alignment database contains large-pose samples~\cite{zhu2012face,jaiswal2013guided,le2012interactive,sagonas2013semi} since labelling standardized landmarks on profile is very challenging. In this section, we demonstrate that labelled profile faces can be well simulated from existing training samples with the help of 3D information.
Inspired by the recent breakthrough in face frontalization~\cite{zhu2015high,hassner2014effective} which generates the frontal view of faces, we propose to invert this process to generate the profile view of faces from medium-pose samples, which is called face profiling. The basic idea is predicting the depth of face image and generating the profile views with 3D rotation.

\subsection{3D Image Meshing}
\begin{figure}[!htb]
  \centering
  \subfigure[]{
  \label{fig-3D-meshing-a}
  \includegraphics[width=0.11\textwidth]{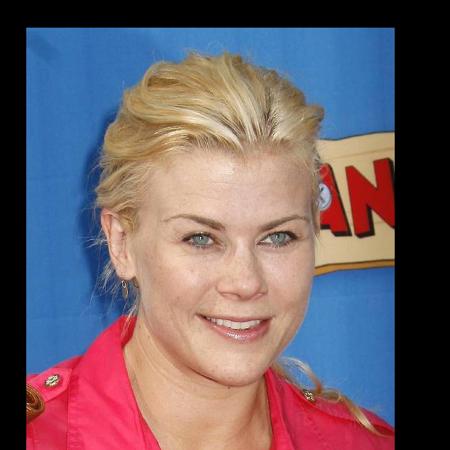}}
  \subfigure[]{
  \label{fig-3D-meshing-b}
  \includegraphics[width=0.11\textwidth]{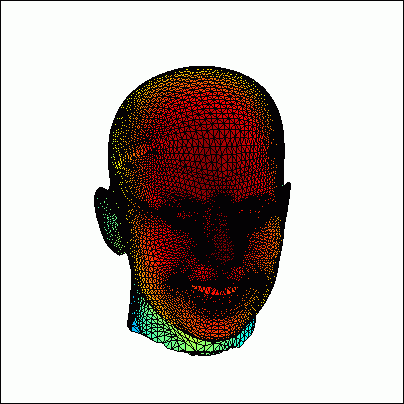}}
  \subfigure[]{
  \label{fig-3D-meshing-c}
  \includegraphics[width=0.11\textwidth]{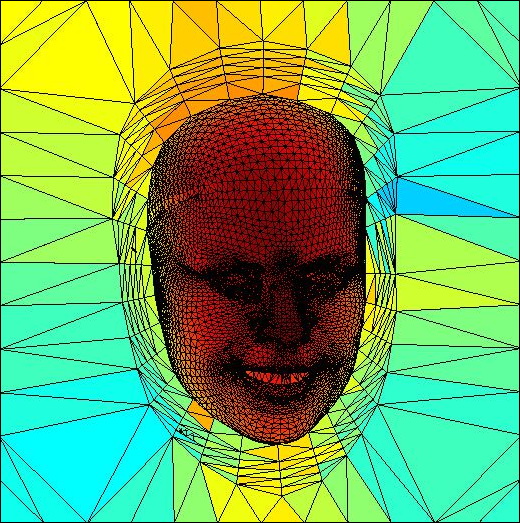}}
  \subfigure[]{
  \label{fig-3D-meshing-d}
  \includegraphics[width=0.11\textwidth]{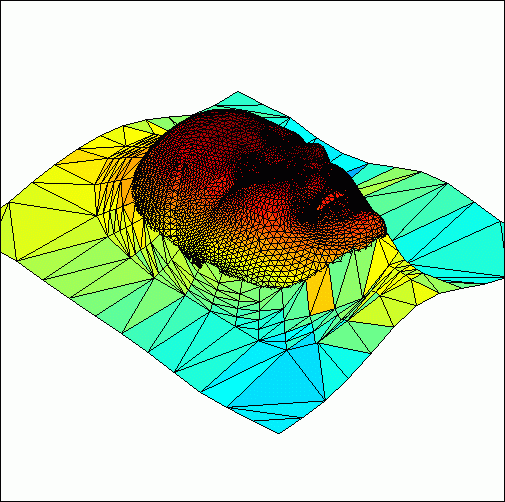}}
  \caption{3D Image Meshing. (a) The input image. (b) The fitted 3D face through MFF. (c) The depth image from 3D meshing. (d) A different view of the depth image.}
  \label{fig-3D-meshing}
\end{figure}

The depth estimation of a face image can be conducted on the face region and external region respectively, with different requirements of accuracy. On the face region, we fit a 3DMM through the Multi-Features Framework~\cite{Romdhani-CVPR-05} (MFF), see Fig.~\ref{fig-3D-meshing-b}. With the ground truth landmarks as a solid constraint throughout the fitting process, the MFF can always converge to a very good result. Few failed samples can be easily adjusted manually. On the external region, we follow the 3D meshing method proposed by Zhu et al.~\cite{zhu2015high} to mark some anchors beyond the face region and estimate their depth, see Fig.~\ref{fig-3D-meshing-c}. Afterwards the whole image is tuned into a 3D object through triangulation, see Fig.~\ref{fig-3D-meshing-c}\ref{fig-3D-meshing-d}.

\subsection{3D Image Rotation}
\begin{figure}[!htb]
  \centering
  \subfigure{
  \includegraphics[width=0.11\textwidth]{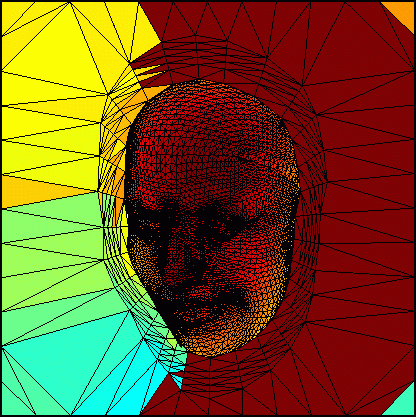}}
  \subfigure{
  \includegraphics[width=0.11\textwidth]{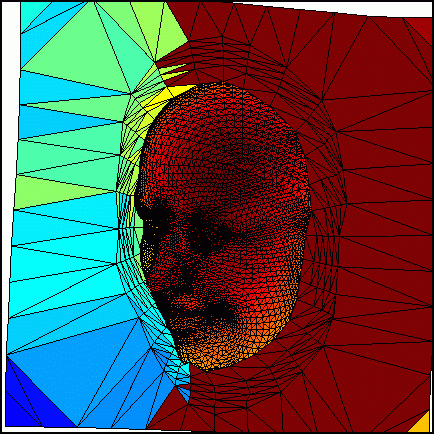}}
  \subfigure{
  \includegraphics[width=0.11\textwidth]{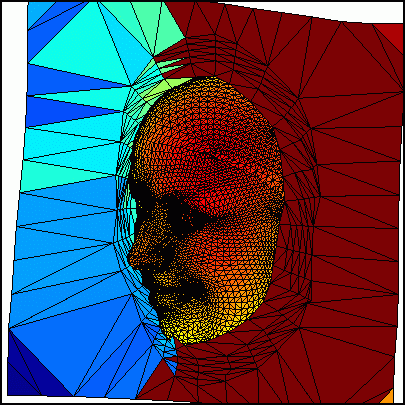}}
  \subfigure{
  \includegraphics[width=0.11\textwidth]{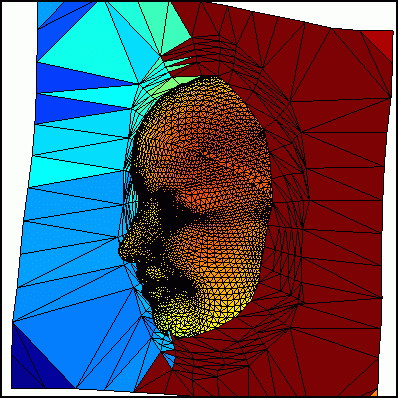}}\\
  \setcounter{subfigure}{0}
  \subfigure[]{
  \label{fig-img-meshing-a}
  \includegraphics[width=0.11\textwidth]{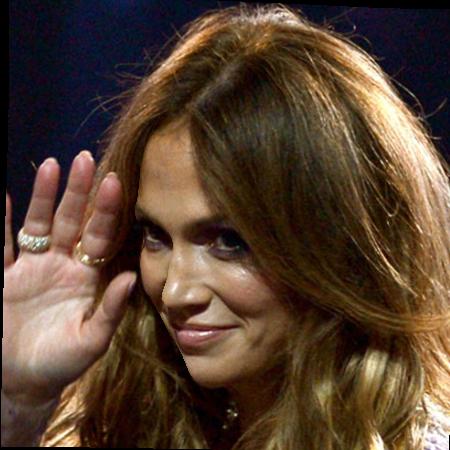}}
  \subfigure[]{
  \label{fig-img-meshing-b}
  \includegraphics[width=0.11\textwidth]{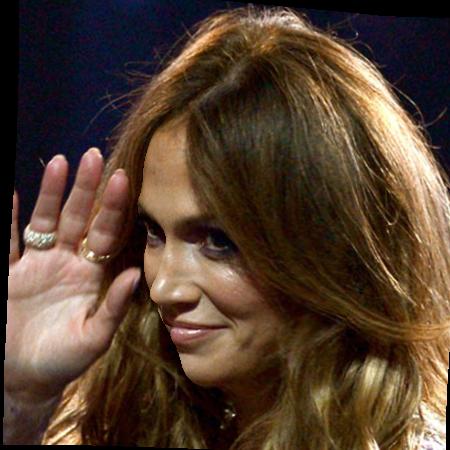}}
  \subfigure[]{
  \label{fig-img-meshing-c}
  \includegraphics[width=0.11\textwidth]{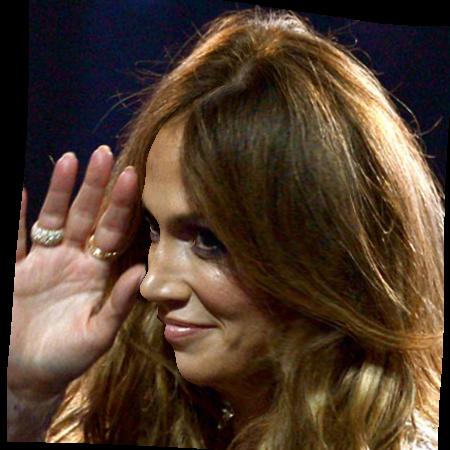}}
  \subfigure[]{
  \label{fig-img-meshing-d}
  \includegraphics[width=0.11\textwidth]{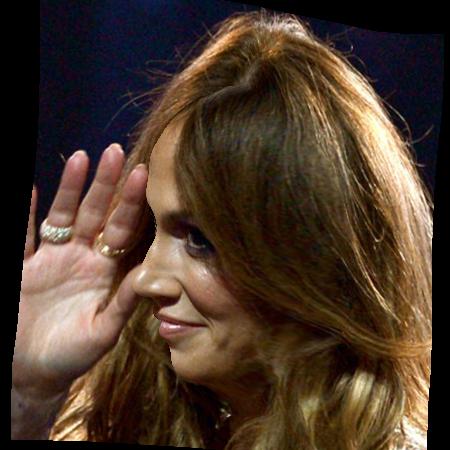}}
  \caption{2D and 3D view of the image rotation. (a) The original yaw angle $yaw_{0}$. (b) $ yaw_{0}+ 20^{\circ}$. (c) $yaw_{0} + 30^{\circ}$. (d) $yaw_{0} + 40^{\circ}$.}
  \label{fig-img-rotation}
\end{figure}

When the depth information is estimated, the face image can be rotated in 3D space to generate the appearances in larger poses (Fig.~\ref{fig-img-rotation}). It can be seen that the external face region is necessary for a realistic profile image. Different from face frontalization, with larger rotation angles the self-occluded region can only be expanded. As a result, we avoid the troubling invisible region filling which may produce large artifacts~\cite{zhu2015high}.

In this work, we enlarge the $yaw$ of the depth image at the step of $5^{\circ}$ until $90^{\circ}$. Through face profiling, we not only obtain the face appearances in large poses and but also augment the dataset to a large scale, which means the CNN can be well trained even given a small database.

\section{Implementation Details}
\subsection{Initialization Regeneration}
With a huge number of parameters, CNN tends to overfit the training set and the networks at deeper cascade might receive training samples with almost zero errors. Therefore we cannot directly adopt the cascade framework as in 2D face alignment. Asthana et al.~\cite{asthana2014incremental} demonstrates that the initializations at each iteration can be well simulated with statistics. In this paper, we also regenerate the $\mathbf{p}^{k}$ but with a more sophisticated method. We observe that the fitting error highly depends on the ground truth face posture (FP). For example, the error of a profile face is mostly caused by a small yaw angle and the error of an open-mouth face is always caused by a close-mouth expression parameter. As a result, it is reasonable to model the perturbation of a training sample with a set of similar-FP samples. In this paper, we define the face posture as the ground truth 2D landmarks without scale and translation:
\begin{equation}\label{equ-model-condition}
  \text{FP}=\mathbf{Pr} * \mathbf{R}^{g}*(\mathbf{\overline{S}} + \mathbf{A}_{id}\bm{\alpha}_{id}^{g} + \mathbf{A}_{exp}\bm{\alpha}_{exp}^{g})_{landmark},
\end{equation}
where $\mathbf{R}^{g},\bm{\alpha}_{id}^{g},\bm{\alpha}_{exp}^{g}$ represent the ground truth pose, shape and expression respectively and the subscript $landmark$ means only landmark points are selected.
Before training, we select two folds of samples as the validation set. For each training sample, we construct a validation subset $\{v_{1},...,v_{m}\}$ whose members share similar FP with the training sample. At iteration $k$, we regenerate the initial parameter by:
\begin{equation}\label{equ-init-regeneration}
  \mathbf{p}^{k} = \mathbf{p}^{g} - (\mathbf{p}^{g}_{v_{i}} - \mathbf{p}^{k}_{v_{i}}),
\end{equation}
where $\mathbf{p}^{k}$ and $\mathbf{p}^{g}$ are the initial and ground truth parameter of a training sample,  $\mathbf{p}^{k}_{v_{i}}$ and $\mathbf{p}^{g}_{v_{i}}$ come from a validation sample $v_{i}$ which is randomly chosen from the corresponding validation subset. Note that $v_{i}$ is never used in training.

\subsection{Landmark Refinement}\label{sec-land-refine}
Dense face alignment method fits all the vertexes of the face model by estimating the model parameters. If we are only interested in a sparse set of points such as landmarks, the error can be further reduced by relaxing the PCA constraint. In the 2D face alignment task, after 3DDFA we extract HOG features at landmarks and train a linear regressor to refine the landmark locations. In fact, 3DDFA can team with any 2D face alignment methods. In the experiment, we also report the results refined by SDM~\cite{Xiong-CVPR-13}.

\section{Experiments}
In this section, we evaluate the performance of 3DDFA in three common face alignment tasks in the wild, i.e., medium-pose face alignment, large-pose face alignment and 3D face alignment. Due to the space constraint, qualitative alignment results are shown in supplemental material.
\subsection{Datasets}\label{sec-datasets}
Evaluations are conducted with three databases, 300W~\cite{sagonas2013300}, AFLW~\cite{kostinger2011annotated} and a specifically constructed AFLW2000-3D database.

\textbf{300W-LP}: 300W~\cite{sagonas2013300} standardises multiple alignment databases with 68 landmarks, including AFW~\cite{zhu2012face}, LFPW~\cite{Belhumeur-2011-LFPW}, HELEN~\cite{zhou2013extensive}, IBUG~\cite{sagonas2013300} and XM2VTS~\cite{messer1999xm2vtsdb}. With 300W, we adopt the proposed face profiling to generate 61,225 samples across large poses (1,786 from IBUG, 5,207 from AFW, 16,556 from LFPW and 37,676 from HELEN, XM2VTS is not used), which is further expanded to 122,450 samples with flipping. We call the database as the 300W across Large Poses (300W-LP)

\textbf{AFLW}: AFLW~\cite{kostinger2011annotated} contains 21,080 in-the-wild faces with large-pose variations (yaw from $-90^{\circ}$ to $90^{\circ}$). Each image is annotated with up to 21 visible landmarks. The dataset is very suitable for evaluating face alignment performance across large poses.

\begin{figure}[!htb]
  \centering
  \includegraphics[width=0.45\textwidth]{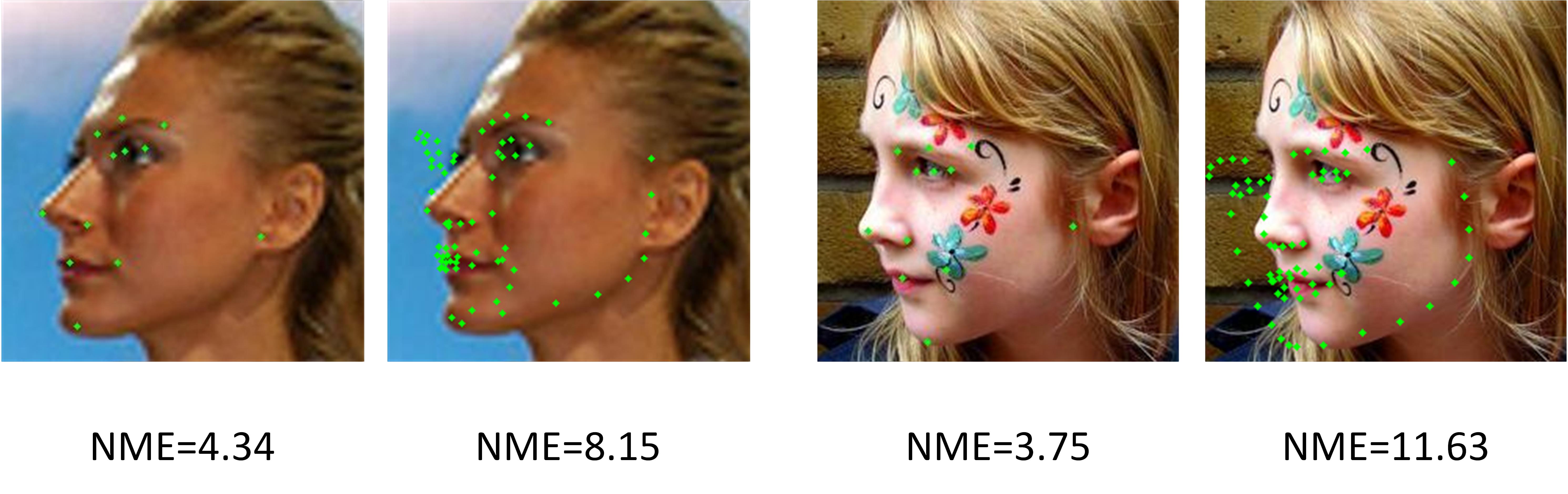}
  \caption{Fake good alignment in AFLW. For each sample, the first shows the visible 21 landmarks and the second shows all the 68 landmarks. The Normalized Mean Error (NME) reflects their accuracy. It can be seen that only evaluating visible landmarks cannot well reflect the fitting accuracy.}
  \label{fig-fakegood}
\end{figure}

\textbf{AFLW2000-3D}: Evaluating 3D face alignment in the wild is difficult due to the lack of pairs of 2D image and 3D model in unconstrained environment. Considering the recent achievements in 3D face reconstruction which can construct a 3D face from 2D landmarks~\cite{Aldrian-PAMI-13,zhu2015high}, we assume that a 3D model can be accurately fitted if sufficient 2D landmarks are provided. Therefore 3D evaluation can be degraded to 2D evaluation which also makes it possible to compare 3DDFA with other 2D face alignment methods. However, AFLW is not suitable for evaluating this task since only visible landmarks lead to serious ambiguity in 3D shape, as reflected by the fake good alignment phenomenon in Fig.~\ref{fig-fakegood}. In this work, we construct a database called AFLW2000-3D for 3D face alignment evaluation, which contains the ground truth 3D faces and the corresponding 68 landmarks of the first 2,000 AFLW samples. Construction details are provided in supplemental material.

\subsection{Performance Analysis}
\begin{figure}[!htb]
  \centering
  \subfigure[]{
  \includegraphics[width=0.23\textwidth]{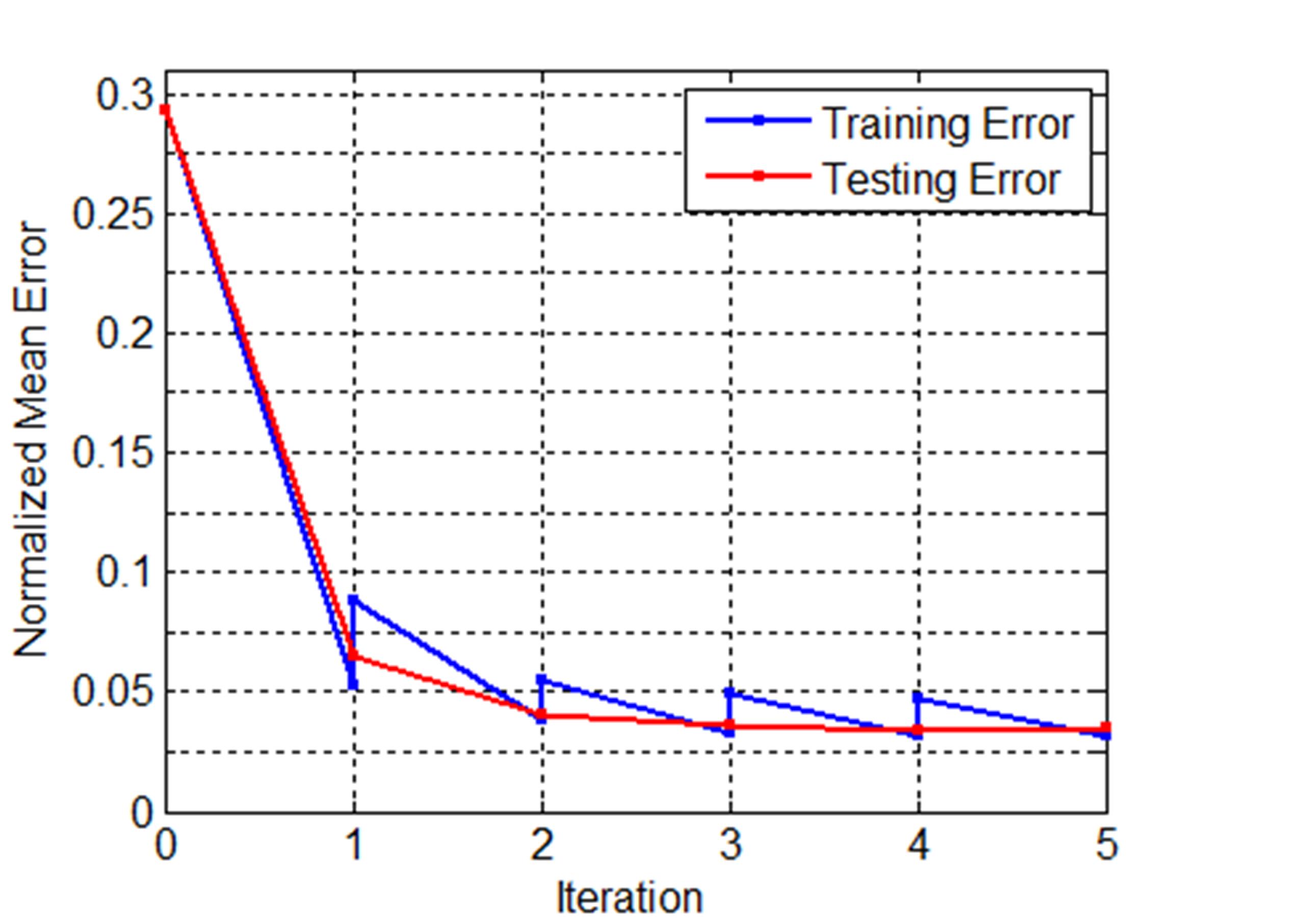}}
  \subfigure[]{
  \includegraphics[width=0.23\textwidth]{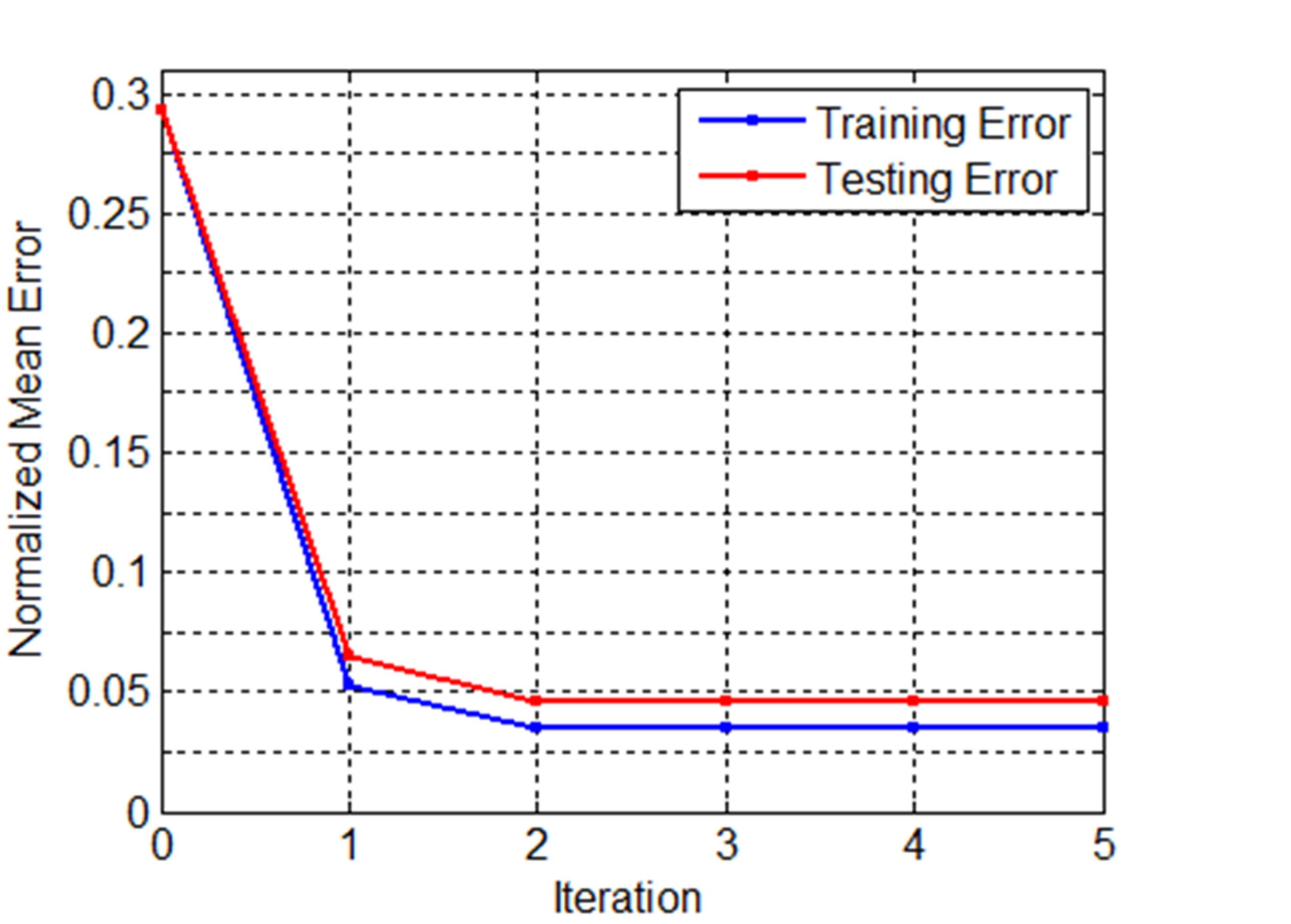}}
  \caption{The training and testing errors with (a) and without (b) initialization regeneration.}
  \label{fig-error-cascade}
\end{figure}

\textbf{Error Reduction in Cascade}: To analyze the error reduction process in cascade and evaluate the effect of initialization regeneration. We divide 300W-LP into 97,967 samples for training and 24,483 samples for testing, without identity overlapping. Fig.~\ref{fig-error-cascade} shows the training and testing errors at each iteration, with and without initialization regeneration.
As observed, the testing error is reduced due to initialization regeneration. In the generic cascade process the training and testing errors converge fast after 2 iterations. While with initialization regeneration, the training error is updated at the beginning of each iteration and the testing error continues to descend.

During testing, 3DDFA takes 25.24ms for each iteration, 17.49ms for PNCC construction on 3.40GHZ CPU and 7.75ms for CNN on GTX TITAN Black GPU. Note that the computing time of PNCC can be greatly reduced if Z-Buffer is conducted on GPU. Considering both effectiveness and efficiency we choose 3 iterations in 3DDFA.

\begin{figure}[!htb]
  \centering
  \includegraphics[width=0.35\textwidth]{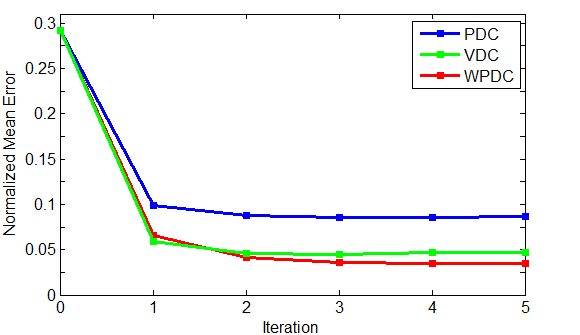}
  \caption{The testing errors with different cost function.}
  \label{fig-error-cost}
\end{figure}

\textbf{Performance with Different Costs}: In this experiment, we demonstrate the performance with different costs including PDC, VDC and WPDC. Fig.~\ref{fig-error-cost} demonstrates the testing errors at each iteration. All the networks are trained until convergence.
It is shown that PDC cannot well model the fitting error and converges to an unsatisfied result. VDC is better than PDC, but the pathological curvature problem makes it only concentrate on a small set of parameters, which limits its performance. WPDC explicitly models the priority of each parameter and adaptively optimizes them with the parameter weights, leading to the best result.

\subsection{Comparison Experiments}
In this paper, we test the performance of 3DDFA on three different tasks, including the large-pose face alignment on AFLW, 3D face alignment on AFLW2000-3D and medium-pose face alignment on 300W.

\begin{table*}[!htb]\small
 \tabcolsep 7pt \caption{The NME(\%) of face alignment results on AFLW and AFLW2000-3D with the first and the second best results highlighted. The bracket shows the training set. The results of provided alignment models are marked with their references.}
  \begin{center}
  \begin{tabular}{| c || c | c | c | c | c || c | c | c | c | c |}
    \hline
     &\multicolumn{5}{c||}{AFLW Dataset (21 pts)} & \multicolumn{5}{c|}{AFLW2000-3D Dataset (68 pts)}\\
    \hline
    \multirow{2}{*}{Method} & \multirow{2}{*}{$[0,30]$} & \multirow{2}{*}{$[30,60]$} & \multirow{2}{*}{$[60,90]$}  & \multirow{2}{*}{Mean} & \multirow{2}{*}{Std} & \multirow{2}{*}{$[0,30]$} & \multirow{2}{*}{$[30,60]$} & \multirow{2}{*}{$[60,90]$}  & \multirow{2}{*}{Mean} & \multirow{2}{*}{Std}\\
    &  & &  &  &  &  &  &  &  & \\
    \hline
    CDM~\cite{yu2013pose} & 8.15 & 13.02 & 16.17 & 12.44 & 4.04 & - & - & - & - & -\\
    \hline
    RCPR~\cite{burgos2013robust} & 6.16 & 18.67 & 34.82 & 19.88 & 14.36 & - & - & - & - & -\\
    RCPR(300W) & 5.40 & 9.80 & 20.61 & 11.94 & 7.83 & 4.16 & 9.88 & 22.58 & 12.21 & 9.43\\
    RCPR(300W-LP) & 5.43 & 6.58 & 11.53 & 7.85 & 3.24 & 4.26 & 5.96 & 13.18 & 7.80 & 4.74\\
    \hline
    ESR(300W) & 5.58 & 10.62 & 20.02 & 12.07 & 7.33 & 4.38 & 10.47 & 20.31 & 11.72 & 8.04\\
    ESR(300W-LP) & 5.66 & 7.12 & 11.94 & 8.24 &  3.29 & 4.60 & 6.70 & 12.67 & 7.99 &  4.19\\
    \hline
    SDM(300W) & \textbf{4.67} & 6.78 & 16.13 & 9.19 & 6.10 & \textbf{3.56} & 7.08 & 17.48 & 9.37 & 7.23\\
    SDM(300W-LP) & \textbf{4.75} & 5.55 & 9.34 & 6.55 & 2.45 & 3.67 & 4.94 & 9.76 & 6.12 & 3.21\\

    \hline
    \hline
    \textbf{3DDFA} & 5.00 & \textbf{5.06} & \textbf{6.74} & \textbf{5.60} & \textbf{0.99} & 3.78 & \textbf{4.54} & \textbf{7.93} & \textbf{5.42} & \textbf{2.21}\\
    \textbf{3DDFA+SDM} & \textbf{4.75} & \textbf{4.83} & \textbf{6.38} & \textbf{5.32} & \textbf{0.92} & \textbf{3.43} & \textbf{4.24} & \textbf{7.17} & \textbf{4.94} & \textbf{1.97} \\
    \hline
  \end{tabular}
  \end{center}
  \label{tab-falp}
\end{table*}

\subsubsection{Large Pose Face Alignment in AFLW}

\textbf{Protocol}: In this experiment, we regard 300W and 300W-LP as the training set respectively and the whole AFLW as the testing set. The bounding boxes provided by AFLW are used for initialization (which are not the ground truth). During training, for 2D methods we use the projected 3D landmarks as the ground truth and for 3DDFA we directly regress the 3DMM parameters. During testing, we divide the testing set into 3 subsets according to their absolute yaw angles: $[0^{\circ},30^{\circ}]$, $[30^{\circ},60^{\circ}]$, and $[60^{\circ},90^{\circ}]$ with 11,596, 5,457 and 4,027 samples respectively. The alignment accuracy is evaluated by the Normalized Mean Error (NME), which is the average of visible landmark error normalised by the bounding box size~\cite{jourabloo2015pose,yu2013pose}. Note that the metric only considers visible landmarks and is normalized by the bounding box size instead of the common inter-pupil distance. Besides, we also report the standard deviation across testing subsets which is a good measure of pose robustness.

\begin{figure}[!htb] \small
  \centering
  \includegraphics[width=0.40\textwidth]{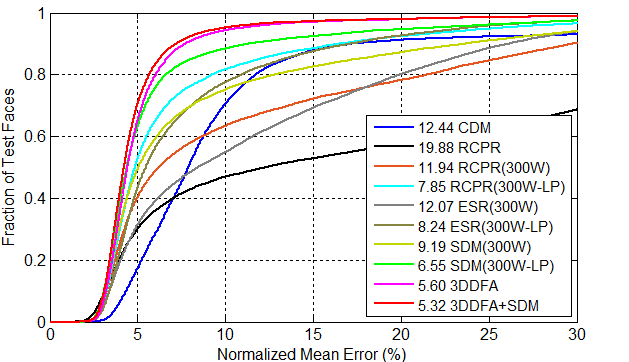}
  \caption{Comparisons of cumulative errors distribution (CED) curves on AFLW. To balance the pose distribution, we plot the CED curves with a subset of 12,081 samples whose absolute yaw angles within $[0^{\circ},30^{\circ}]$, $[30^{\circ},60^{\circ}]$ and $[60^{\circ},90^{\circ}]$ are 1/3 each.}
  \label{fig-ced-aflw}
\end{figure}

\textbf{Methods}: Since little experiment has been conducted on AFLW, we choose some baseline methods with released codes, including CDM~\cite{yu2013pose}, RCPR~\cite{burgos2013robust}, ESR~\cite{Cao-CVPR-12} and SDM~\cite{yan2013learn}. Among them ESR and SDM are popular face alignment methods in recent years. CDM is the first one claimed to perform pose-free face alignment. RCPR is a occlusion-robust method with the potential to deal with self-occlusion and we train it with landmark visibility labels computed by~\cite{hassner2014effective}. Table.~\ref{tab-falp} demonstrates the comparison results and Fig.~\ref{fig-ced-aflw} shows the corresponding CED curves. Each method is trained on 300W and 300W-LP respectively to demonstrate the boost from face profiling. If a trained model is provided in the code, we also demonstrate its performance. Since CDM only contains testing code, we just report its performance with the provided alignment model. For 3DDFA which depends on large scales of data, we only report its performance trained on 300W-LP, with the network structure in Fig.~\ref{fig-overview}.

\textbf{Results}: Firstly, the results indicate that all the methods benefits substantially from face profiling when dealing with large poses. The improvements in $[60^{\circ},90^{\circ}]$ are $44.06\%$ for RCPR, $40.36\%$ for ESR and $42.10\%$ for SDM. This is especially impressive since the alignment models are trained on the synthesized data and tested on real samples. Thus the fidelity of the face profiling method can be well demonstrated. Secondly, 3DDFA reaches the state of the art above all the 2D methods especially beyond medium poses. The minimum standard deviation of 3DDFA also demonstrates its robustness to pose variations. Finally, the performance of 3DDFA can be further improved with the SDM landmark refinement in Section~\ref{sec-land-refine}.

\subsubsection{3D Face Alignment in AFLW2000-3D}
\begin{figure}[!htb] \small
  \centering
  \includegraphics[width=0.40\textwidth]{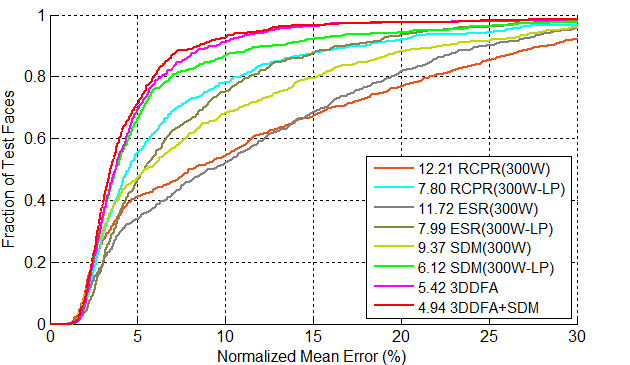}
  \caption{Comparisons of cumulative errors distribution (CED) curves on AFLW2000. To balance the pose distribution, we plot the CED curves with a subset of 696 samples whose absolute yaw angles within $[0^{\circ},30^{\circ}]$, $[30^{\circ},60^{\circ}]$ and $[60^{\circ},90^{\circ}]$ are 1/3 each.}
  \label{fig-ced-aflw2000}
\end{figure}

As described in Section~\ref{sec-datasets}, 3D face alignment evaluation can be degraded to all-landmark evaluation considering both visible and invisible ones. Using AFLW2000-3D as the testing set, this experiment follows the same protocol as AFLW, except 1) Instead of the visible 21 landmarks, all the MultiPIE-68 landmarks~\cite{sagonas2013300} in AFLW2000-3D are used for evaluation. 2) With the ground truth 3D models, the ground truth bounding boxes enclosing all the landmarks are provided for initialization. There are 1,306 samples in $[0^{\circ},30^{\circ}]$, 462 samples in $[30^{\circ},60^{\circ}]$ and 232 samples in $[60^{\circ},90^{\circ}]$. The results are demonstrates in Table.~\ref{tab-falp} and the CED curves are plot in Fig.~\ref{fig-ced-aflw2000}. We do not report the performance of provided CDM and RCPR models since they do not detect invisible landmarks.

Compared with the results in AFLW, we can see the defect of barely evaluating visible landmarks. For all the methods, despite with ground truth bounding boxes the performance in $[60^{\circ},90^{\circ}]$ and the standard deviation are obviously reduced when considering all the landmarks. We think for 3D face alignment which depends on both visible and invisible landmarks~\cite{Aldrian-PAMI-13,jeni2015dense}, evaluating all the landmarks are necessary.

\subsubsection{Medium Pose Face Alignment}
Even though not aimed at advancing face alignment in medium poses, we are also interested in the performance of 3DDFA in this popular task. The experiments are conducted on 300W following the common protocol in~\cite{zhu2015face}, where we use the training part of
LFPW, HELEN and the whole AFW for training (3,148 images and 50,521 after augmentation), and perform testing on three parts: the test samples from LFPW and HELEN as the common subset, the 135-image IBUG as the challenging subset, and the union of them as the full set (689 images in total). The alignment accuracy are evaluated by standard landmark mean error normalised by the inter-pupil distance (NME).
\begin{table}[!htb]\small
 \tabcolsep 8pt \caption{The NME(\%) of face alignment results on 300W, with the first and the second best results highlighted.}
  \begin{center}
  \begin{tabular}{ c  c  c  c }
    \hline
    Method & Common & Challenging & Full \\
    \hline
    TSPM~\cite{zhu2012face} & 8.22 & 18.33 & 10.20\\

    ESR~\cite{Cao-CVPR-12} & 5.28 & 17.00 & 7.58  \\

    RCPR~\cite{burgos2013robust} & 6.18 & 17.26 & 8.35 \\

    SDM~\cite{Xiong-CVPR-13} & 5.57 & 15.40 & 7.50 \\

    LBF~\cite{ren2014face} & \textbf{4.95} & 11.98 & 6.32 \\

    CFSS~\cite{zhu2015face} & \textbf{4.73} & \textbf{9.98} & \textbf{5.76} \\
    \hline
    \textbf{3DDFA} & 6.15 & 10.59 & 7.01 \\
    \textbf{3DDFA+SDM} & 5.53 & \textbf{9.56} & \textbf{6.31}\\
    \hline
  \end{tabular}
  \end{center}
  \label{tab-fa-medium}
\end{table}
It can be seen in Tabel.~\ref{tab-fa-medium} that even as a generic face alignment algorithm, 3DDFA still demonstrates competitive performance on the common set and state-of-the-art performance on the challenging set.

\section{Conclusions}
In this paper, we propose a novel method, 3D Dense Face Alignment (3DDFA), which well solves the problem of face alignment across large poses. Different from the traditional landmark detection framework, 3DDFA fits a dense 3D morphable model with cascaded CNN to solve the self-occlusion in modelling and the high nonlinearity in fitting in large poses. We also propose a face profiling algorithm to synthesize face appearances in profile view, which can provide abundant samples for training. Experiments show the state-of-the-art performance in AFLW, AFLW2000-3D and 300W. In future work, we believe that 3DDFA can be further improved with more complicated network architecture, like larger input size and deeper network.

{\small
\bibliographystyle{ieee}
\bibliography{reference_zxy}
}

\end{document}